 \documentclass[final,5p,times,twocolumn,authoryear]{elsarticle}

\usepackage{url}
\usepackage{cite}
\usepackage{amsmath,amssymb,amsfonts}
\usepackage{algorithmic}
\usepackage{graphicx}
\usepackage{float}
\usepackage{textcomp}
\usepackage{colortbl} 
\usepackage{xcolor}
\usepackage[colorlinks = true,
            linkcolor = red,
            urlcolor  = red,
            citecolor = blue,
            anchorcolor = blue]{hyperref}
\usepackage{multirow}
\usepackage{tabularray}
\usepackage{booktabs}
\usepackage{rotating}
\usepackage{subcaption}
\usepackage{tabularx,ragged2e}
\usepackage{wrapfig}

\newcommand{\tablestyle}[2]{\setlength{\tabcolsep}{#1}\renewcommand{\arraystretch}{#2}\centering\footnotesize}
\journal{Medical Image Analysis}

\begin{document}

\begin{frontmatter}



\title{MedCLIP-SAMv2: Towards Universal Text-Driven Medical Image Segmentation} 

\author[label1]{Taha Koleilat, Hojat Asgariandehkordi, Hassan Rivaz}
\author[label2]{Yiming Xiao}
\affiliation{organization={Department of Electrical and Computer Engineering, Concordia University},
            city={Montreal},
            state={Quebec},
            country={Canada}}

\affiliation{organization={Department of Computer Science and Software Engineering, Concordia University},
            city={Montreal},
            state={Quebec},
            country={Canada}}

\begin{abstract}
Segmentation of anatomical structures and pathologies in medical images is essential for modern disease diagnosis, clinical research, and treatment planning. While significant advancements have been made in deep learning-based segmentation techniques, many of these methods still suffer from limitations in data efficiency, generalizability, and interactivity. As a result, developing robust segmentation methods that require fewer labeled datasets remains a critical challenge in medical image analysis. Recently, the introduction of foundation models like CLIP and Segment-Anything-Model (SAM), with robust cross-domain representations, has paved the way for interactive and universal image segmentation. However, further exploration of these models for data-efficient segmentation in medical imaging is an active field of research. In this paper, we introduce MedCLIP-SAMv2, a novel framework that integrates the CLIP and SAM models to perform segmentation on clinical scans using text prompts, in both zero-shot and weakly supervised settings. Our approach includes fine-tuning the BiomedCLIP model with a new Decoupled Hard Negative Noise Contrastive Estimation (DHN-NCE) loss, and leveraging the Multi-modal Information Bottleneck (M2IB) to create visual prompts for generating segmentation masks with SAM in the zero-shot setting. We also investigate using zero-shot segmentation labels in a weakly supervised paradigm to enhance segmentation quality further. Extensive validation across four diverse segmentation tasks and medical imaging modalities (breast tumor ultrasound, brain tumor MRI, lung X-ray, and lung CT) demonstrates the high accuracy of our proposed framework. Our code is available at \url{https://github.com/HealthX-Lab/MedCLIP-SAMv2}.
\end{abstract}




\end{frontmatter}



\section{Introduction}
\label{introduction}
With the growing availability of radiological technologies, there is an increasing demand for precise and efficient medical image segmentation to support the study, diagnosis, and treatment of various medical conditions \citep{siuly2016medical}. Deep learning (DL) techniques have emerged as state-of-the-art (SOTA) in this field; however, they face three key challenges that hinder their broader clinical adoption. First, the scarcity of large, well-annotated datasets presents a major obstacle to DL model development. Second, the lack of interactivity and interpretability undermines trust in these methods. Finally, most medical DL models are trained for specific tasks and contrasts/modalities, limiting their flexibility. While several self-supervised and weakly supervised approaches \citep{baevski2023efficient,chen2020big,taleb2021multimodal} have been introduced to improve training efficiency, and explainable AI (XAI) techniques, including uncertainty estimation \citep{loquercio2020general,liu2020simple} and saliency maps \citep{arun2021assessing,bae2020rethinking} are under active investigation, cross-domain generalization remains a major challenge.

Recently, the introduction of foundation models, such as Contrastive Language-Image Pre-Training (CLIP) \citep{radford2021learning} and Segment Anything Model (SAM) \citep{kirillov2023segment} have paved the way for interactive and universal medical image segmentation. Several research groups have adapted CLIP and SAM for radiological tasks, including the development of BiomedCLIP \citep{zhang2023largescale} and MedSAM \citep{Ma2023SegmentAI}, which were pre-trained on vast amounts of biomedical data. However, further advances in parameter fine-tuning methods could enhance the performance of these models in radiology.

Although CLIP training primarily operates at a global level for image-text mapping, research \citep{fu2024featup} has revealed that these models can encode rich feature representations of images. This allows us to establish the relationship between global textual information and local visual features \citep{zhou2022extract, rao2022denseclip}, which can be exploited for efficient zero-shot medical image segmentation, enabling broader applicability even in data-scarce settings, as we explored for the first time in our MICCAI 2024 paper \citep{koleilat2024medclipsambridgingtextimage}. The complex and nuanced nature of medical terminology, combined with the subtle and intricate variations inherent in medical images, introduces unique challenges that are less pronounced in natural images. While adapting CLIP to the medical image domain may seem attractive, it is non-trivial and requires substantial ground truth labels to fine-tune the model effectively, especially for segmentation downstream tasks \citep{poudel2023exploring}. The lack of large, high-quality annotated datasets in medical imaging further exacerbates this challenge. This calls for biomedical domain-specific CLIP models, such as BiomedCLIP \citep{zhang2023largescale} and effective fine-tuning loss functions based on these domain-specific CLIP models to establish more effective cross-modal learning in radiological applications, such as pathology localization, segmentation, and diagnosis. We continue to explore these in this paper for MedSAM-CLIPv2.

On the other hand, as interest in SAM grows, to mitigate its reliance on visual prompts (e.g., point and/or bounding box) for segmentation, which require prior clinical knowledge, recent methods have emerged to fine-tune SAM without these prompts \citep{chen2024unsam,hu2023efficiently}, generate prompts via Class Activation Maps (CAM) from classification tasks \citep{li2024clipsam,li2023clip,liu2024weakly}, and refine its output using weak supervision \citep{yang2023foundation,chen2023segment,huang2023push}. While still in its early stages, the use of foundation models for interactive and universal medical image segmentation remains an important area for further exploration. Recently, to address these challenges, we introduced MedCLIP-SAM in MICCAI2024 \citep{koleilat2024medclipsambridgingtextimage}, which leverages BiomedCLIP \citep{zhang2023largescale} to generate text-based box prompts for SAM \citep{kirillov2023segment} towards interactive and universal medical image segmentation, in both zero-shot and weakly supervised settings. Following the preliminary success, further improvement and exploration of the framework are necessary to further elevate the performance and gain deeper insights into the CLIP and SAM foundation models for medical imaging applications. As a result, in this paper, we propose MedCLIP-SAMv2, a novel technique that further evolves and significantly improves upon our original MedCLIP-SAM framework for zero-shot and weakly supervised medical image segmentation \citep{koleilat2024medclipsambridgingtextimage}. Specifically, the prominent upgrades for the newly proposed MedCLIP-SAMv2 framework from the original method include:
\begin{itemize}
    \item We investigated different saliency map generation techniques for CLIP models, where we replaced gScoreCAM \citep{chen2022gScoreCAM} with M2IB \citep{m2ib}, which, when combined with our fine-tuning of BiomedCLIP \citep{zhang2023largescale}, significantly improved zero-shot segmentation accuracy.
    \item We enhanced weakly supervised segmentation results and interpretability from the previous framework by training nnUNet \citep{isensee2021nnu} using pseudo-labels while providing uncertainty estimation via checkpoint ensembling \citep{zhao2022efficient}.
    \item The validation was expanded by incorporating an additional Lung CT dataset, thereby covering four key radiological modalities — CT, MRI, ultrasound, and X-ray. This comprehensive testing further demonstrates the framework's versatility and robustness across diverse segmentation tasks.
    \item We investigated and optimized advanced text prompt engineering strategies by leveraging large language model (LLM) reasoning and various ensembling methods, which are shown to significantly boost zero-shot segmentation performance.
    \item Significantly more extensive experiments were conducted for further validation of our framework's design components, including testing different SAM backbones and visual prompt types. We meticulously evaluate the necessity of each component in our framework and demonstrate their individual contribution to the overall performance enhancement.
\end{itemize}

The newly proposed MedCLIP-SAMv2 framework is more accurate, advancing further toward universal text-driven medical image segmentation with an increase of 13.07\% and 11.21\% in Dice score for zero-shot and weakly supervised paradigms, respectively. Our main contributions are threefold: \textbf{First}, we propose a new CLIP training/fine-tuning loss function called Decoupled Hard Negative Noise Contrastive Estimation (DHN-NCE). \textbf{Second}, we introduce a text-driven zero-shot medical segmentation method, combining CLIP and SAM for radiological tasks. \textbf{Lastly}, we explore a weakly-supervised strategy to further refine zero-shot segmentation results with uncertainty estimation. Our proposed framework is extensively validated across four distinct segmentation tasks and modalities, including breast tumor segmentation in ultrasound, brain tumor segmentation in MRI, and lung segmentation in chest X-ray and CT.

\section{Related Work}
\subsection{CLIP in Medical Domain} Several works have utilized CLIP for medical images and texts. Despite being trained on 400 million natural image-text pairs, CLIP's performance suffers on medical tasks. For this reason, works like PubMedCLIP \citep{eslami-etal-2023-pubmedclip} suggested fine-tuning CLIP on a set of PubMed articles for the task of Medical Question-Answering; Zhang \textit{et al.} \citep{zhang2023largescale} later showed PubMedCLIP's poor performance on cross-modal retrieval tasks (worse than CLIP). On the other hand, MedCLIP \citep{wang2022medclip} proposed a technique to utilize decoupled images and texts in training to augment data while Windsor \textit{et al.} \citep{windsor2023visionlanguage} explored different methods of enhancing the performance of vision-language models for medical domain tasks in a limited data setting. Alternatively, Wu \textit{et al.} \citep{wu2023medklip} proposed a method of enhancing the text in medical reports by simplifying the sentence complexity. Moreover, other works like \citep{keicher2023flexr} and \citep{Tiu2022} have utilized CLIP for the task of pathology detection and medical report generation. However, notably, almost all mentioned works \citep{wang2022medclip,windsor2023visionlanguage,wu2023medklip,keicher2023flexr,Tiu2022} only utilized Chest X-ray data for their proposed methods. BiomedCLIP \citep{zhang2023largescale} is by far the most recent work for multi-modal medical data on a large scale, which was shown to be superior for cross-modal retrieval accuracy. Notable studies \citep{koleilat2024biomedcoop, poudel2023exploring} have investigated the transfer capabilities of BiomedCLIP in downstream tasks such as classification and segmentation. However, its adaptability remains largely unexplored compared to the extensive body of CLIP literature. To the best of our knowledge, our work is the first to explore the potential of BiomedCLIP in zero-shot segmentation tasks, paving the way for more efficient usage in medical imaging.

\subsection{Weakly Supervised Semantic Segmentation}
To mitigate the shortage of well-labeled datasets for medical image segmentation, many works have explored weakly supervised paradigms for segmenting distinct regions in natural images with CLIP-like models. CLIP-ES \citep{lin2023clip} proposed a purely text-driven approach to producing better pseudo-masks through CLIP's class activation maps instead of training affinity networks, while SAMS \citep{yang2023foundation} later extended the work by making use of the SAM model to produce coarse and fine seeds from image-level labels. Additionally, SG-WSSS \citep{jiang2023segment} studied different visual prompting methods, including scribbles, points, and bounding boxes to prompt SAM through CAM scores. However, these works may fail to translate well to medical scans, which have different characteristics than natural images. Novel CAM techniques specifically tailored for CLIP models like gScoreCAM \citep{chen2022gScoreCAM} and M2IB \citep{m2ib} have emerged with SOTA performance for generating multi-modal saliency maps. Specifically, gScoreCAM \citep{chen2022gScoreCAM} utilized the top-K channel activations from the text and image encoder layers, leading to better-localized saliency maps. The more recent M2IB \citep{m2ib} reformulates the information bottleneck theory to multi-modal applications, where it was proven to outperform CAM-based, perturbation-based, and attention-based saliency mapping techniques. Additionally, M2IB also demonstrated its potential for medical image applications, where a fine-tuned CLIP model on Chest X-ray datasets was shown to properly highlight regions of abnormalities \citep{m2ib}. Recently, Liu \textit{et al.} \citep{liu2023chatgpt} focus on improving interpretation of zero-shot medical image diagnosis through engineering relevant text prompts by integrating ChatGPT that outputs relevant descriptions of the radiological abnormality. However, these previous works don't inspect improving medical segmentation through model training.

\subsection{SAM for Medical Imaging Segmentation}
With the advent of SAM, a foundation model for image segmentation that enables zero-shot generalization through a promptable architecture consisting of a powerful image encoder, a flexible prompt encoder, and a lightweight mask decoder, a myriad of research has been dedicated to adapting it for medical imaging purposes. MedSAM \citep{Ma2023SegmentAI} provided a large-scale fine-tuning of SAM on about 1 million medical image-mask pairs and demonstrated superior performance when it comes to multiple segmentation tasks. AutoSAM \citep{Shaharabany2023AutoSAMAS} offered a more efficient approach to fine-tuning SAM on medical images through training the prompt encoder and developing a lightweight deconvolution mask decoder for medical segmentation tasks. Cheng \textit{et al.} \citep{Cheng2023SAMOM} found that bounding boxes gave the best results when prompting SAM across 12 different medical segmentation tasks, and Huang \textit{et al.} \citep{huang2023push} proposed a pseudo-mask correction framework to enhance noisy labels generated from SAM for medical images that can be used for further fine-tuning. Finally, Gong \textit{et al.} \citep{gong20233dsamadapter} replaced SAM's mask decoder with a 3D convolutional neural network so that volumetric medical images can be supported.
\section{Methods}
\label{section:methods}
\begin{figure}[h]
\includegraphics[width=0.5\textwidth]{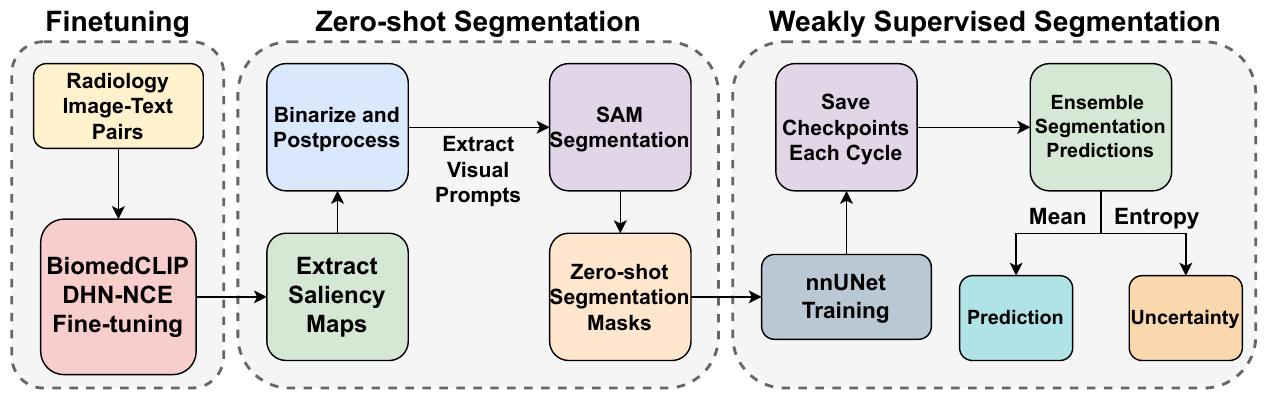}
\caption{A general overview of the essential components.} 
\label{fig:General Figure}
\end{figure}

\begin{figure*}[t!]
\includegraphics[width=\textwidth]{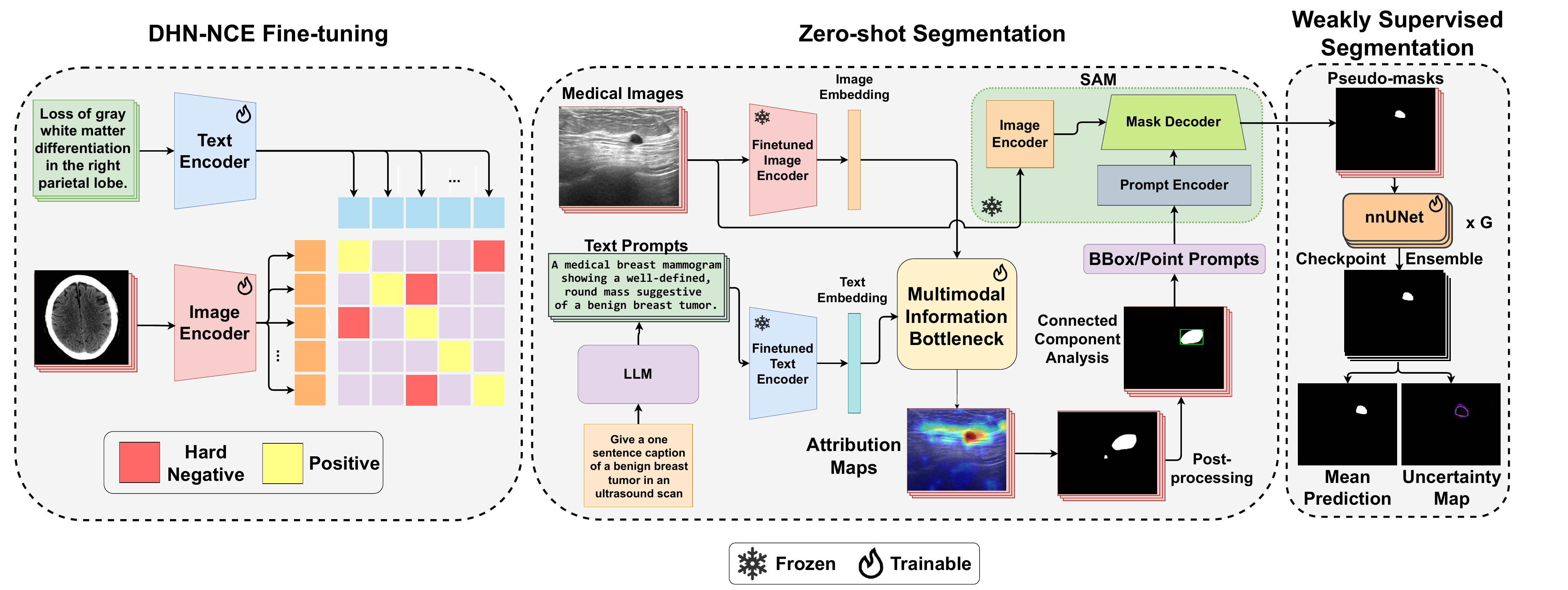}
\caption{An overview of the proposed MedCLIP-SAMv2 framework.} 
\label{fig:System Figure}
\end{figure*}

A full overview of the proposed MedCLIP-SAMv2 framework is presented in Fig. \ref{fig:System Figure}, which is organized into three distinct stages:  1) BiomedCLIP fine-tuning employing our new DHN-NCE loss, 2) zero-shot segmentation guided by text-prompts, and 3) weakly supervised segmentation for potential label refinement. We additionally showcase a summary of the main components of the framework in Fig. \ref{fig:General Figure} for the readers' easy reference.
\begin{figure}
    \centering
    \includegraphics[width=0.5\textwidth,height=0.3\textwidth]{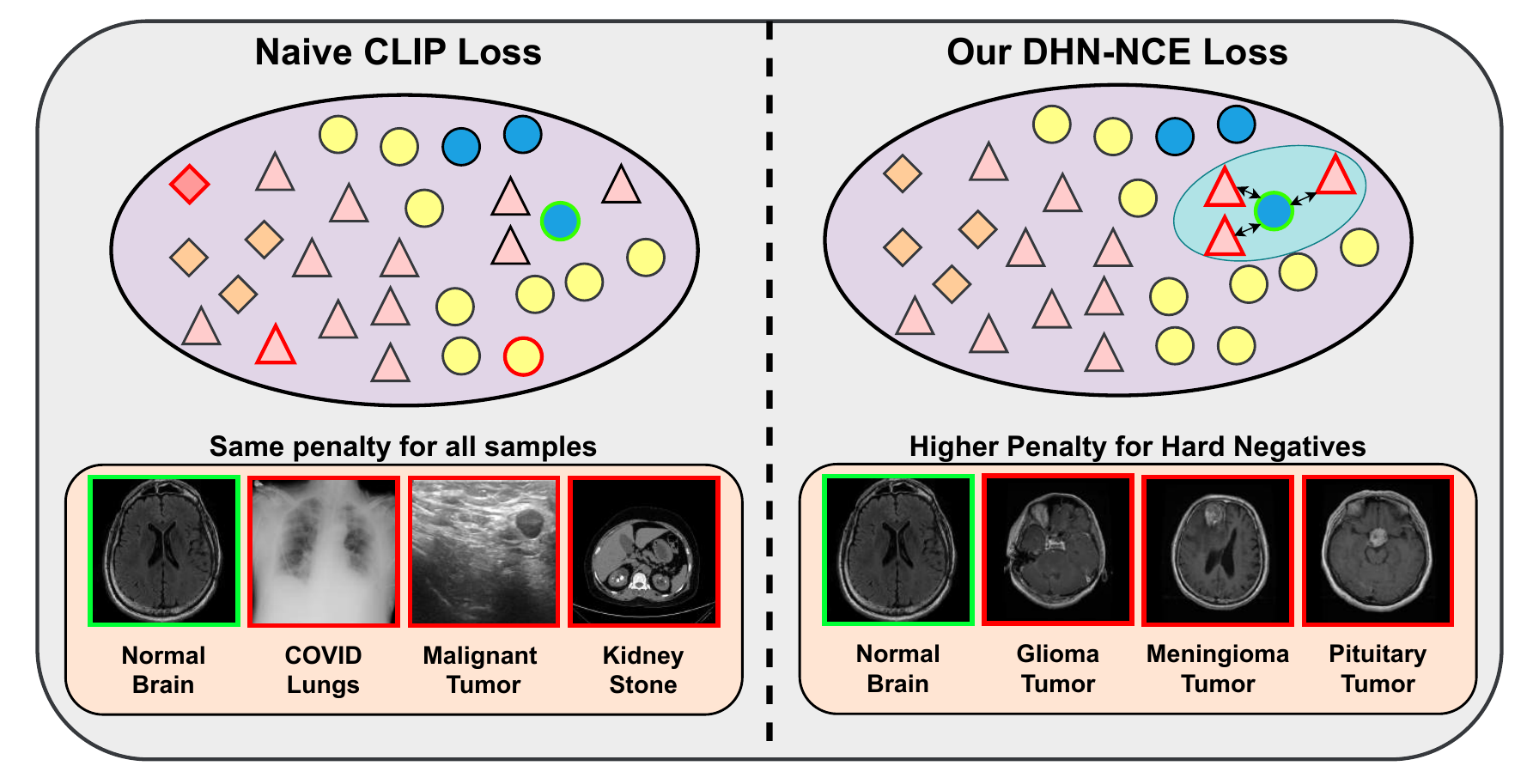}
    \caption{Comparison of the standard CLIP loss, which applies uniform penalties to all examples regardless of difficulty, with our DHN-NCE loss, which prioritizes harder examples. The DHN-NCE loss enhances the differentiation of medical cases by appropriately penalizing close negatives through adaptive weighting formulas. Green outline represents the anchor example while the red outline represents the negative examples.}
    \label{fig:loss-illustration}
\end{figure}
\subsection{Efficient DHN-NCE Fine-tuning}
CLIP-like models are commonly trained on extensive datasets consisting of images paired with their corresponding textual descriptions. These models employ an image encoder and a text encoder to extract features, representing them as vectors in a shared dimensional space\footnote{It is important to note that CLIP utilizes the global [CLS] tokens of the final vision and text encoder layers before projection to the shared embedding space}: $\mathbf{I}_{p,i}$ for images and $\mathbf{T}_{p,i}$ for texts. Through the mechanism of contrastive learning, CLIP aligns semantically related image-text pairs by minimizing their distance in the embedding space while maximizing the separation of unrelated pairs. This shared embedding framework facilitates a cohesive understanding of multimodal data.  Although BiomedCLIP \citep{zhang2023largescale} was trained on medical charts/images and clinical texts, further fine-tuning can significantly enhance its performance on tasks specific to medical imaging. In traditional CLIP training with the InfoNCE loss \citep{oord2018representation}, the \textit{negative-positive-coupling (NPC)} effect \citep{yeh2022decoupled} can reduce learning efficiency, especially with smaller batch sizes. Additionally, for medical images, distinguishing between subtle differences in cases within the same imaging category can be challenging. To address these issues, we propose the Decoupled Hard Negative Noise Contrastive Estimation (DHN-NCE) loss, which 1) combines the InfoNCE loss \citep{oord2018representation} with hard negative sampling \citep{robinson2021contrastive}, emphasizing ``close samples", and 2) incorporates decoupled contrastive learning \citep{yeh2022decoupled} by removing the positive term in the denominator, allowing for smaller batch sizes. \\
\\
\textbf{Original InfoNCE Loss:} The standard InfoNCE loss for contrastive learning is formulated as follows:

\begin{equation}
    \mathcal{L}_{\text{InfoNCE}} = -\sum_{i=1}^{B} \log \frac{\exp(\mathbf{z}_i^\top \mathbf{z}_i^+ / \tau)}{\sum_{j=1}^{B} \exp(\mathbf{z}_i^\top \mathbf{z}_j / \tau)}
\end{equation}

\noindent
where \( B \) is the batch size, \(\mathbf{z}_i\) represents the feature embedding of the anchor sample, \(\mathbf{z}_i^+\) is the positive pair for \( \mathbf{z}_i \), \( \tau \) is the temperature parameter, and \( B \) is the batch size.
\\
\\
\textbf{InfoNCE for Vision-Language Matching:} To get a vision-language contrastive loss, we replace generic embeddings with image (\( \mathbf{I}_{p,i} \)) and text (\( \mathbf{T}_{p,i} \)) embeddings. In this context, $t \rightarrow v$ refers to text-to-image, while $v \rightarrow t$ indicates image-to-text:

\begin{equation}
    \mathcal{L}^{v \rightarrow t} = -\sum_{i=1}^{B} \log \frac{\exp(\mathbf{I}_{p,i}^\top \mathbf{T}_{p,i} / \tau)}{\sum_{j=1}^{B} \exp(\mathbf{I}_{p,i}^\top \mathbf{T}_{p,j} / \tau)}
\end{equation}

\begin{equation}
    \mathcal{L}^{t \rightarrow v} = -\sum_{i=1}^{B} \log \frac{\exp(\mathbf{T}_{p,i}^\top \mathbf{I}_{p,i} / \tau)}{\sum_{j=1}^{B} \exp(\mathbf{T}_{p,i}^\top \mathbf{I}_{p,j} / \tau)}
\end{equation}
\\
\\
\textbf{Decoupling Positives and Negatives:} Expanding the logarithm and separating terms of Eq (2) and (3), we obtain:

\begin{equation}
    \mathcal{L}^{v \rightarrow t} = -\sum_{i=1}^{B} \left[ \frac{\mathbf{I}_{p,i}^\top \mathbf{T}_{p,i}}{\tau} - \log \sum_{j=1}^{B} \exp(\mathbf{I}_{p,i}^\top \mathbf{T}_{p,j} / \tau) \right]
\end{equation}

\begin{equation}
    \mathcal{L}^{t \rightarrow v} = -\sum_{i=1}^{B} \left[ \frac{\mathbf{T}_{p,i}^\top \mathbf{I}_{p,i}}{\tau} - \log \sum_{j=1}^{B} \exp(\mathbf{T}_{p,i}^\top \mathbf{I}_{p,j} / \tau) \right]
\end{equation}
\noindent

Since the summation in the denominators of Eq (2) and (3) includes both the positive pair (\( j = i \)) and the negatives (\( j \neq i \)), we split it as:

\begin{equation}
    \label{eqn:6}
    \sum_{j=1}^{B} \exp(\mathbf{I}_{p,i}^\top \mathbf{T}_{p,j} / \tau)=\exp(\mathbf{I}_{p,i}^\top \mathbf{T}_{p,i} / \tau) + \sum_{j \neq i} \exp(\mathbf{I}_{p,i}^\top \mathbf{T}_{p,j} / \tau)
\end{equation}

\noindent
Following the positive-negative decoupling approach in \citep{yeh2022decoupled}, we remove the positive pair and obtain the decoupled vision-language contrastive loss:

\begin{equation}
    \mathcal{L}^{v \rightarrow t} = -\sum_{i=1}^{B} \frac{\mathbf{I}_{p,i}^\top \mathbf{T}_{p,i}}{\tau} + \sum_{i=1}^{B} \log \left( \sum_{j \neq i} e^{\mathbf{I}_{p,i}^\top \mathbf{T}_{p,j} / \tau} \right)
\end{equation}

\begin{equation}
    \mathcal{L}^{t \rightarrow v} = -\sum_{i=1}^{B} \frac{\mathbf{T}_{p,i}^\top \mathbf{I}_{p,i}}{\tau} + \sum_{i=1}^{B} \log \left( \sum_{j \neq i} e^{\mathbf{T}_{p,i}^\top \mathbf{I}_{p,j} / \tau} \right)
\end{equation}
\\
\\
\textbf{Applying Hardness Weights:} The resulting DHN-NCE loss function, $\mathcal{L}_{DHN-NCE}$, employs weighting functions ($\mathcal{W}_{\mathbf{I}_{p,i}\mathbf{T}_{p,j}}^{v \rightarrow t}$, $\mathcal{W}_{\mathbf{T}_{p,i}\mathbf{I}_{p,j}}^{t \rightarrow v}$) to increase the penalty for negative samples that are close to the anchor, using image-to-text and text-to-image hardness parameters $\beta_1, \beta_2 \geq 0$.
\\
\\
\begin{equation}
    \resizebox{0.89\hsize}{!}{%
        $\mathcal{L}^{v \rightarrow t} = -\sum\limits_{i=1}^B \frac{\mathbf{I}_{p,i} \mathbf{T}_{p,i}^\top}{\tau} +  \sum\limits_{i=1}^B \log\left(\sum\limits_{j \neq i} e^{\mathbf{I}_{p,i} \mathbf{T}_{p,j}^\top/\tau} \mathcal{W}_{\mathbf{I}_{p,i} \mathbf{T}_{p,j}}^{v \rightarrow t}\right)$%
        }
\end{equation}
\begin{equation}
    \resizebox{0.89\hsize}{!}{%
        $\mathcal{L}^{t \rightarrow v} = -\sum\limits_{i=1}^B\frac{\mathbf{T}_{p,i}\mathbf{I}_{p,i}^\top}{\tau} + \sum\limits_{i=1}^Blog\left(\sum\limits_{j \neq i}e^{\mathbf{T}_{p,i}\mathbf{I}_{p,j}^\top/\tau}\mathcal{W}_{\mathbf{T}_{p,i}\mathbf{I}_{p,j}}^{t \rightarrow v}\right)$%
        }
\end{equation}
\begin{equation}
    \mathcal{L}_{DHN-NCE} = \mathcal{L}^{v \rightarrow t} + \mathcal{L}^{t \rightarrow v} 
\end{equation}
where the hardness weighting formulas are as follows:
\begin{equation}
    \mathcal{W}_{\mathbf{I}_{p,i}\mathbf{T}_{p,j}}^{v \rightarrow t} = (B-1)\times\frac{e^{\beta_1\mathbf{I}_{p,i}\mathbf{T}_{p,j}/\tau}}{\sum_{k \neq i}e^{\beta_1\mathbf{I}_{p,i}\mathbf{T}_{p,k}/\tau}}
\end{equation}
\begin{equation}
    \mathcal{W}_{\mathbf{T}_{p,i}\mathbf{I}_{p,j}}^{t \rightarrow v} = (B-1)\times\frac{e^{\beta_2\mathbf{T}_{p,i}\mathbf{I}_{p,j}/\tau}}{\sum_{k \neq i}e^{\beta_2\mathbf{T}_{p,i}\mathbf{I}_{p,k}/\tau}}
\end{equation} 
The weighting functions leverage exponential scaling to amplify the contributions of hard negatives—those with higher similarity scores—while suppressing easier negatives, ensuring the total weight distribution prioritizes these challenging cases (see Fig \ref{fig:loss-illustration}). By decoupling the positive term from the denominator, DHN-NCE prevents easy positives from diminishing the gradients associated with hard negatives. This mechanism sharpens the model’s focus on refining distinctions for harder cases, enabling more efficient training even with small batch sizes. Such properties make DHN-NCE particularly suited for medical imaging tasks with limited data availability and subtle feature variations.

\subsection{Zero-shot Medical Image Segmentation}
In this stage, we utilize the fine-tuned BiomedCLIP with the updated parameters \(\boldsymbol{\theta} = \{\boldsymbol{\theta}_\text{img}, \boldsymbol{\theta}_\text{text}\}\) as the backbone model for feature extraction from both images and text prompts. The core segmentation process relies on the Multi-modal Information Bottleneck (M2IB) technique \citep{m2ib}, which generates visual saliency maps of the target regions by associating text prompts with image regions.
\\
\\
The zero-shot segmentation pipeline can be described as follows:
\\
\\
\textbf{Image and Text Embedding Extraction}: Given input medical images \( \mathbf{I} \) and their corresponding text prompts \( \mathbf{T} \), the image encoder \(\Phi_{\text{img}} \) and the text encoder \(\Phi_{\text{text}}\) from the fine-tuned BiomedCLIP model are used to extract embeddings:
\begin{equation}
    \mathbf{Z}_\text{img} = \Phi_{\text{img}}(\mathbf{I};\boldsymbol{\theta}_\text{img}) 
\end{equation}
\begin{equation}
    \mathbf{Z}_\text{text} = \Phi_{\text{text}}(\mathbf{T};\boldsymbol{\theta}_\text{text})
\end{equation}
\\
\textbf{LLM Prompt Generation:}
Since BiomedCLIP utilizes text captions from PubMed to pre-train its text encoder (i.e., PubMedBERT), we utilize LLMs like GPT-4 \citep{achiam2023gpt} to generate sophisticated text prompts that can guide the model to localize certain salient regions. Specifically, we generate descriptive textual prompts that can guide the model to pay attention to salient features in the medical scan as follows: \\
\texttt{Give a sentence caption that describes unique visual features of [TARGET] in [MODALITY]} \\
We can engineer different prompt configurations ranging from generic to class-specific context captions and we study the effect of these different styles in Section \ref{sec:text-prompt-ablation}.
\\
\\
\noindent \textbf{Saliency Map Generation}: The embeddings \( \mathbf{Z}_\text{img} \) and \( \mathbf{Z}_\text{text} \) are then passed through the Multi-modal Information Bottleneck (M2IB) module \citep{m2ib}, which learns to align the image and text modalities by maximizing the mutual information between the input image $\mathbf{I}$ and a good representative text prompt $\mathbf{T}$ while filtering out irrelevant information between the image embedding and the input image. By doing so, the process bridges the semantic gap between encoded visual and textual features ensuring that embeddings emphasize the content that is jointly relevant across image and text. Specifically, the M2IB module introduces a stochastic information bottleneck $\boldsymbol{\lambda}_S$ $\in$ $\mathbb{R}^{H \times W}$ such that \(0 \leq \boldsymbol{\lambda}_S \leq 1\) where $H$,$W$ are the respective height and weight of the input image $\mathbf{I}$. This produces a continuous visual saliency map for the image representing the importance of each pixel concerning the text prompt. This visual saliency map is produced by optimizing the following objective function:
\begin{equation}
    \boldsymbol{\lambda_S} = MI(\mathbf{Z}_\text{img},\mathbf{Z}_\text{text}; \boldsymbol{\theta}) - \gamma \times MI(\mathbf{Z}_\text{img}, \mathbf{I}; \boldsymbol{\theta})
\end{equation}
\noindent
where \(MI\) is the mutual information operation and $\gamma$ is a hyperparameter that balances the trade-off between
relevance and compression.
\\
\\
\textbf{Post-processing for Initial Segmentation}: 
To obtain a discrete pixel-wise segmentation, we apply Otsu's thresholding \citep{Otsu} to the saliency map \(\boldsymbol{\lambda_S} \), automatically determining an optimal threshold \(\eta^*\) that separates foreground (regions of interest) from background by minimizing intra-class variance. The binarized segmentation is then given by:
\begin{equation}
    \mathbf{Y}_\text{otsu} = \begin{cases} 
    1, & \boldsymbol{\lambda}_S(x, y) \geq \eta^* \\ 
    0, & \boldsymbol{\lambda}_S(x, y) < \eta^*
    \end{cases}
\end{equation}
After thresholding, small, disconnected contours may still exist. To refine the segmentation and ensure robust results, we perform Connected Component Analysis on the identified contours $C$. For each connected component \( c \in C \), we compute a confidence score based on the saliency map \( \boldsymbol{\lambda_S} \). The confidence of a connected component is derived as follows:

\begin{equation}
    \text{Confidence}(c) = \frac{\sum\limits_{i \in c} p_i \cdot \hat{y}_i}{\sum\limits_{i \in c} \hat{y}_i},
\end{equation}

\noindent
where \(p_i\) is the probability that pixel \(i\) belongs to the foreground class, and \(\hat{y}_i\) is the predicted binary label for pixel \(i\). Using this confidence score, we identify the most confident connected components to form the final coarse segmentation:

\begin{equation}
    \mathbf{Y}_\text{coarse} = \{ c \in C : \text{Confidence}(c) > \eta_c \},
\end{equation}

\noindent
where \(\eta_c\) is a confidence threshold. This process refines the initial segmentation by removing unreliable regions and retaining only high-confidence contours.
\\
\\
\textbf{Segmentation Refinement via SAM}: The initial segmentation is used as input to the Segment Anything Model, which refines the segmentation by taking visual prompts \(\mathbf{V}\) (e.g., bounding boxes or points) derived from the post-processed clusters. For bounding boxes, we calculate 4 box coordinates (bounding boxes) that enclose each connected contour in the coarse segmentation, while for points, we randomly sample different points that lie within the contour. The final zero-shot segmentation mask \( \mathbf{Y}_\text{zero-shot} \) is thus produced as:

\begin{equation}
   \mathbf{Y}_\text{zero-shot} = \textit{SAM}(\mathbf{Y}_\text{coarse}; \mathbf{V}) 
\end{equation}
\noindent

\subsection{Uncertainty-Aware Weakly Supervised Segmentation}
To further enhance the segmentation accuracy, the zero-shot segmentation results \( \mathbf{Y}_\text{zero-shot} \) are then used as pseudo-labels with the input medical images \(\mathbf{I}\) to train a segmentation network \(\mathbf{M}\) in a weakly supervised manner. Thus, the training data will be \(\mathcal{T}\) = \( \{(\mathbf{I}, \mathbf{Y}_\text{zero-shot})\}\). Building on the recent work by Zhao \textit{et al.} \citep{zhao2022efficient}, checkpoint ensembling has demonstrated superior effectiveness in uncertainty estimation for medical image segmentation when compared to techniques such as Monte Carlo Dropout and mean-field Bayesian Neural Networks. This finding is particularly relevant in the context of the nnUNet framework \citep{isensee2021nnu}. Given a total number of epochs \(E\), the training process is divided into \( D \) cycles composed of \(E_d = \frac{E}{D}\) epochs, and during each cycle, we save \( G_d \) checkpoints of the model. Importantly, this checkpoint strategy adds no delays to the training process, as it involves saving alternate checkpoints of the same model rather than training separate models. After completing all training cycles, the probabilistic prediction of the final segmentation \( \mathbf{Y}_\text{final} \) is obtained by averaging the predictions from the \(G = D * G_d\) total checkpoints saved during the training process providing a Monte-Carlo-like approximation:
\begin{equation}
    p(\mathbf{Y}_\text{final}\vert \mathbf{X}; \mathcal{T}) \approx \frac{1}{G} \sum_{n=1}^{G} p(\mathbf{Y}_\text{final}| \mathbf{X}; \mathbf{M}_{n})
\end{equation}
where \( \mathbf{M}_{n} \) represents the weights of the model at the \( n \)-th checkpoint, and \( \mathbf{X} \) are unseen testing input images. 
\\
\\
\textbf{Segmentation Uncertainty Estimation}:
The variation in predictions across different checkpoints also allows for estimating uncertainty in the final segmentation mask. The generated uncertainty map helps pinpoint regions of the medical scan that exhibit high uncertainty in the prediction. Given \(R\) classes in the medical image, the uncertainty for each pixel \((i,j)\) can be computed by calculating the entropy as follows: 
\begin{equation}
    H(\mathbf{Y}_{\text{final},{(i,j)}}) = -\sum_{r=1}^R h(r) \log h(r)
    \label{eq:bayesian_uncertainty_ent}
\end{equation}
where
\begin{equation}
    h(r) = p(\mathbf{Y}_{\text{final}, (i,j)} = r \vert \mathbf{X}; \mathcal{T})
\end{equation}
\subsection{Datasets and Experimental Setup}
\subsubsection{BiomedCLIP fine-tuning} 
We employed the public MedPix \citep{siragusa2024medpix} dataset, which contains various radiological modalities, to fine-tune the BiomedCLIP model \citep{zhang2023largescale} with our DHN-NCE loss. The base encoders for images and text were the Vision Transformer (ViT) and PubMedBERT \citep{zhang2023largescale}, respectively. The MedPix dataset was cleaned by removing special characters, trimming leading and trailing white spaces, and excluding samples with captions shorter than 20 characters. All images were resized to 224 $\times$ 224 pixels and normalized according to the RGB channel means and standard deviations used in the original CLIP model \citep{radford2021learning}. We performed an 85\%-15\% split, resulting in 20,292 training images and 3,515 validation images. Fine-tuning was performed with a learning rate of 1E-6, a 50\% decay rate, and a batch size of 64.

To validate the fine-tuning quality of BiomedCLIP, we assessed the top-1 and top-2 accuracy of matching retrievals for both image-to-text and text-to-image on the ROCO (Radiology Objects in COntext) dataset \citep{Pelka2018RadiologyOI}, which contains approximately 7,042 multi-modal medical images covering a wide range of radiological cases. We ran the experiments five times with a batch size of 50, using shuffling to randomize image-text pairs (resulting in 70,420 shuffled examples). In addition, we compared different SOTA loss functions for fine-tuning, including InfoNCE \citep{oord2018representation}, DCL \citep{yeh2022decoupled} and HN-NCE \citep{rdk+23} against our DHN-NCE loss. For a fair comparison, all strategies were trained using the same hyperparameters ($\tau$ = 0.6, learning rate = 1E-6), with the hardness parameters for HN-NCE and DHN-NCE set to $\beta_1$ = $\beta_2$ = 0.15. As a reference, we also included baseline results from pre-trained BiomedCLIP \citep{zhang2023largescale}, PMC-CLIP \citep{lin2023pmcclip}, and CLIP \citep{radford2021learning}.
\noindent
\subsubsection{Datasets}
To evaluate the zero-shot and weakly supervised segmentation results, as well as various design elements of the proposed MedCLIP-SAMv2 framework, we utilized four public datasets, each representing different radiology modalities and tasks. These datasets, which include segmentations of breast tumors, brain tumors, and lungs, were divided into training, validation, and testing sets as follows: 
\begin{itemize}
    \item \textbf{Breast Tumor Ultrasound}: The Breast Ultrasound Images dataset (BUSI) \citep{ALDHABYANI2020104863}, containing 600 images of benign and malignant tumors for training. Additionally, 50 and 113 images from the UDIAT dataset \citep{Byra2020-gg} were used for validation and testing, respectively.
    \item \textbf{Brain Tumor MRI}: The Brain Tumor dataset \citep{Cheng2017}, comprising 1,462 T1-weighted MRI scans for training, 1,002 for validation, and 600 for testing. 
    \item \textbf{Lung Chest X-ray}: The COVID-19 Radiography Database (COVID-QU-Ex) \citep{9144185,RAHMAN2021104319} is divided into 16,280 chest X-rays (normal, lung opacity, viral pneumonia, and COVID-19 cases) for training, 1,372 for validation, and 957 for testing.
    \item \textbf{Lung CT}: CT scans from \citep{Konya2020-ix}, consisting of segmentation masks for fibrotic diseased lungs from 107 patients, split into 7,959 slices for training, 3,010 for validation, and 1,800 for testing. The split was done by patient ID to prevent data leakage.
\end{itemize}

\begin{table*}[t!]
\centering
\tablestyle{-27pt}{1.1}
\addtolength{\tabcolsep}{+30pt}
\renewcommand{\arraystretch}{1.5}  
\resizebox{\textwidth}{!}{%
\begin{tabular}{cccccccccccc}
\toprule
\multirow{2}{*}{\textbf{Technique}} & \multirow{2}{*}{\textbf{Method}} & \multicolumn{2}{c}{\textbf{Breast Ultrasound}}                                      & \multicolumn{2}{c}{\textbf{Brain MRI}}                                              & \multicolumn{2}{c}{\textbf{Lung X-ray}}                                             & \multicolumn{2}{c}{\textbf{Lung CT}}                                                & \multicolumn{2}{c}{\textbf{All}}                                \\ \cmidrule{3-12} 
                                    &                                  & \textbf{DSC} $\uparrow$ & \textbf{NSD} $\uparrow$ & \textbf{DSC} $\uparrow$ & \textbf{NSD} $\uparrow$ & \textbf{DSC} $\uparrow$ & \textbf{NSD} $\uparrow$ & \textbf{DSC} $\uparrow$ & \textbf{NSD} $\uparrow$ & \textbf{DSC} $\uparrow$ & \textbf{NSD} $\uparrow$ \\ \hline
\multirow{3}{*}{Zero-shot}          & SaLIP   & 44.33$_{10.12}$                                    & 48.62$_{10.25}$                                    & 47.96$_{9.14}$                                    & 50.24$_{9.26}$                                    & 63.14$_{11.34}$                                    & 66.44$_{11.58}$                                    & 76.32$_{11.22}$                                    & 78.46$_{11.35}$                                    & 57.94$_{10.49}$                                    & 60.94$_{10.65}$                                  \\
                                    & SAMAug      & 56.39$_{10.85}$                                    & 59.23$_{10.92}$                                    & 45.71$_{10.34}$                                    & 48.81$_{11.29}$                                    & 57.18$_{12.12}$                                    & 60.08$_{12.34}$                                    & 44.61$_{10.42}$                                    & 46.48$_{10.57}$                                    & 50.97$_{10.96}$                                    & 53.65$_{11.30}$                                   \\
                                    & MedCLIP-SAM     & 67.82$_{8.26}$   & 69.12$_{9.12}$   & 66.72$_{5.27}$   & 68.01$_{6.16}$   & 64.49$_{9.09}$   & 65.89$_{10.44}$  & 59.14$_{9.52}$   & 60.47$_{9.98}$   & 64.54$_{8.20}$   & 66.10$_{9.08}$ \\
                                    & \cellcolor[gray]{0.92}Ours                                                                      & \cellcolor[gray]{0.92}{77.76$_{9.52}$}                              & \cellcolor[gray]{0.92}{81.11$_{9.89}$}                              & \cellcolor[gray]{0.92}{76.52$_{7.06}$}                           & \cellcolor[gray]{0.92}{82.23$_{7.13}$}                           & \cellcolor[gray]{0.92} 75.79$_{3.44}$                           & \cellcolor[gray]{0.92} 80.88$_{3.52}$                         & \cellcolor[gray]{0.92} 80.38$_{5.81}$                          & \cellcolor[gray]{0.92} 82.03$_{5.94}$                         & \cellcolor[gray]{0.92} {77.61$_{6.82}$}                           & \cellcolor[gray]{0.92} {81.56$_{7.00}$}                           \\ \midrule
\multirow{2}{*}{Weakly Supervised}                  & nnUNet                                                                   &  73.77$_{14.48}$                       &      79.71$_{14.79}$                      &        77.16$_{12.17}$                               &       85.21$_{12.60}$                              &    70.15$_{6.40}$                                  &  74.10$_{6.59}$                                      & 82.24$_{5.12}$                                     &     85.65$_{4.70}$                                    &  75.83$_{10.31}$                       &  81.17$_{10.52}$                            \\
& MedCLIP-SAM     & 58.62$_{5.66}$   & 60.94$_{5.87}$   & 58.80$_{8.63}$   & 61.77$_{8.64}$   & \textbf{86.07$_{8.61}$}   & \textbf{88.65$_{8.09}$}  & 80.12$_{8.38}$   & 83.73$_{8.29}$   & 70.90$_{7.92}$   & 73.77$_{7.80}$ \\
&   \cellcolor[gray]{0.83} Ours                                                                  & \cellcolor[gray]{0.83} \cellcolor[gray]{0.83} \textbf{78.87$_{12.29}$}                           & \cellcolor[gray]{0.83} \textbf{84.58$_{12.19}$}                           &   \cellcolor[gray]{0.83} \textbf{80.03$_{9.91}$}                                      & \cellcolor[gray]{0.83}  \cellcolor[gray]{0.83} \textbf{88.25$_{10.04}$}                                       &  \cellcolor[gray]{0.83} 80.77$_{4.44}$                                        &   \cellcolor[gray]{0.83} 84.53$_{4.51}$                                      &       \cellcolor[gray]{0.83}     \cellcolor[gray]{0.83} \textbf{88.78$_{4.43}$}                              & \cellcolor[gray]{0.83} \textbf{91.95$_{4.06}$}                                          &  \cellcolor[gray]{0.83}  \textbf{82.11$_{8.49}$}                                     & \cellcolor[gray]{0.83} \textbf{87.33$_{8.46}$}                                      \\ \midrule
\multirow{2}{*}{One-shot}           & UniverSeg    
& 40.56$_{5.14}$                                    & 53.25$_{6.22}$                                    & 23.81$_{5.45}$                                    & 35.28$_{6.49}$                                    & 68.15$_{2.21}$                                    & 80.09$_{2.16}$                                    & 54.94$_{8.21}$                                    & 69.62$_{7.59}$                                    & 46.87$_{5.67}$                                    & 59.56$_{5.98}$                                    \\

                                    & ProtoSAM                      &
                                    48.44$_{10.93}$                                    & 50.24$_{10.84}$                                    & 45.68$_{15.14}$                                    & 51.69$_{15.65}$                                    & 80.75$_{1.40}$                                    & 85.11$_{1.30}$                                    & 84.50$_{9.94}$                                     & 87.62$_{9.72}$                                    & 64.84$_{10.60}$                                    & 68.67$_{10.71}$                                    \\ \midrule
\multirow{2}{*}{Few-shot (K = 4)}   & UniverSeg                    & 47.56$_{8.57}$                                    & 54.25$_{8.71}$                                    & 53.82$_{10.17}$                                    & 66.40$_{9.96}$                                    & 79.25$_{2.10}$                                    & 84.80$_{1.70}$                                    & 65.68$_{12.02}$                                    & 70.56$_{11.67}$                                    & 61.58$_{9.02}$                                    & 69.00$_{8.86}$                                    \\
                                    & Self-Prompt-SAM               & 42.04$_{17.19}$                                    & 44.30$_{17.64}$                                    & 46.43$_{15.25}$                                    & 50.29$_{15.83}$                                    & 67.97$_{2.89}$                                    & 71.63$_{2.83}$                                    & 81.50$_{3.84}$                                    & 83.40$_{3.77}$                                    & 59.49$_{11.74}$                                    & 62.41$_{12.08}$                                    \\ \midrule
\multirow{2}{*}{Few-shot (K = 16)}  & UniverSeg                       & 66.36$_{8.57}$                                    & 72.22$_{8.30}$                                    & 62.82$_{7.97}$                                    & 72.76$_{7.94}$                                    & 83.44$_{1.54}$                                    & 87.73$_{1.24}$                                    & {86.49$_{2.49}$}                                    & {89.96$_{1.94}$}                                    & 74.78$_{6.03}$                                    & 80.67$_{5.86}$                                    \\
                                    & Self-Prompt-SAM              & 62.36$_{16.38}$                                    & 66.01$_{16.92}$                                    & 52.55$_{15.29}$                                    & 57.07$_{15.93}$                                    & {82.49$_{2.50}$}                                    & {86.49$_{2.45}$}                                    & 83.66$_{3.90}$                                    & 85.49$_{3.84}$                                    & 70.27$_{11.44}$                                    & 73.77$_{11.84}$                                    \\ \midrule[0.3pt]\midrule[0.3pt]
\multirow{2}{*}{Fully Supervised}                   & nnUNet                                                                     &  82.47$_{10.49}$                         &  88.32$_{10.77}$                       &  87.74$_{6.28}$                                        &        95.10$_{6.28}$                                 &            98.72$_{0.65}$                                 &  99.51$_{0.41}$                                   &  97.10$_{2.74}$                          &  99.18$_{2.13}$                                      &  84.63$_{6.27}$                                      &  90.42$_{6.33}$                                      \\
                                    & nnUNet Ensemble  
                                    & 84.72$_{10.97}$                               & 90.85$_{11.26}$                                   & 88.82$_{5.93}$                                   & 95.84$_{5.54}$                              & 99.14$_{2.50}$                              & 99.82$_{1.93}$                           & 98.12$_{4.09}$                              & 99.65$_{4.03}$                                   & 85.43$_{6.68}$                                & 91.74$_{6.66}$
                                   \\ \bottomrule
\end{tabular}
}
\caption{Comparison of DSC and NSD values (\%) with different few-shot and zero-shot medical image segmentation methods (mean$_{std}$)}
\label{tab:sota_results}
\end{table*}
\begin{table*}[t!]
\centering
\tablestyle{-19pt}{1.1}
\addtolength{\tabcolsep}{+30pt}
\resizebox{\textwidth}{!}{%
\begin{tabular}{cccccc}
\toprule
\multirow{2}{*}{Model} & \multirow{2}{*}{Version} & \multicolumn{2}{c}{$image \rightarrow text 
 $ (\%)}                         & \multicolumn{2}{c}{$text \rightarrow image 
 $ (\%)}                         \\ \cmidrule{3-6} 
                       &                          & \multicolumn{1}{c}{Top-1}          & Top-2          & \multicolumn{1}{c}{Top-1}          & Top-2          \\ \midrule
CLIP \citep{radford2021learning}                   & Pre-trained              & 26.68$_{0.30}$ & 41.80$_{0.19}$ & 26.17$_{0.20}$ & 41.13$_{0.20}$ \\ \midrule
PMC-CLIP \citep{lin2023pmcclip}            & Pre-trained              & \multicolumn{1}{c}{75.47$_{0.37}$} & 87.46$_{0.11}$ & \multicolumn{1}{c}{76.78$_{0.11}$} & 88.35$_{0.19}$ \\ \midrule
\multirow{5}{*}{BiomedCLIP \citep{zhang2023largescale}} &
  Pre-trained &
  \multicolumn{1}{c}{81.83$_{0.20}$} &
  92.79$_{0.13}$ &
  \multicolumn{1}{c}{81.36$_{0.48}$} &
  92.27$_{0.14}$ \\
                       & InfoNCE \citep{oord2018representation}                 & \multicolumn{1}{c}{84.21$_{0.35}$} & 94.47$_{0.19}$ & \multicolumn{1}{c}{85.73$_{0.19}$} & 94.99$_{0.16}$ \\
                       & DCL \citep{yeh2022decoupled}                   & \multicolumn{1}{c}{84.44$_{0.37}$} & 94.68$_{0.19}$ & \multicolumn{1}{c}{85.89$_{0.16}$} & 95.09$_{0.19}$ \\
                       & HN-NCE \citep{rdk+23}                 & \multicolumn{1}{c}{84.33$_{0.35}$} & 94.60$_{0.19}$ & \multicolumn{1}{c}{85.80$_{0.17}$} & 95.10$_{0.19}$ \\
 &
  \cellcolor[gray]{0.9} \textbf{DHN-NCE (ours)} &
  \cellcolor[gray]{0.9} \textbf{84.70$_{0.33}$} & 
 \cellcolor[gray]{0.9} \textbf{94.73$_{0.16}$} & 
  \cellcolor[gray]{0.9} \textbf{85.99$_{0.19}$} & \cellcolor[gray]{0.9} \textbf{95.17$_{0.19}$} \\ \bottomrule
\end{tabular}
}
\caption{Top-K cross-modal retrieval accuracy (mean$_{std}$) 
for CLIP models.}
\label{tab:cross-modal}
\end{table*}
\begin{table*}[htb]
\centering
\tablestyle{-24pt}{1.1}
\addtolength{\tabcolsep}{+30pt}
\resizebox{\textwidth}{!}{%
\begin{tabular}{ccccccccc}
\toprule
\multirow{2}{*}{\textbf{Prompt}} & \multicolumn{2}{c}{\textbf{Breast Ultrasound}}                                                                       & \multicolumn{2}{c}{\textbf{Brain MRI}}                                                                               & \multicolumn{2}{c}{\textbf{Lung X-ray}}                                                                              & \multicolumn{2}{c}{\textbf{Lung CT}}                                                                                 \\ \cmidrule{2-9} 
                                 & \textbf{DSC} $\uparrow$ & \textbf{NSD} $\uparrow$ & \textbf{DSC} $\uparrow$ & \textbf{NSD} $\uparrow$ & \textbf{DSC} $\uparrow$ & \textbf{NSD} $\uparrow$ & \textbf{DSC} $\uparrow$ & \textbf{NSD} $\uparrow$ \\ \midrule

\textbf{P0}                               & 63.79$_{15.12}$                                                     & 67.89$_{15.08}$                                                     & 70.98$_{7.61}$                                                     & 76.42$_{7.63}$                                                     & \textbf{75.79$_{3.44}$}                                            & \textbf{80.88$_{3.52}$}                                            & 69.89$_{5.14}$                                                     & 71.83$_{4.98}$                                                     \\ \midrule
\textbf{P1}                               & 67.66$_{14.35}$                                                   & 71.56$_{14.78}$                                                     & 37.19$_{10.98}$                                                     & 39.77$_{11.63}$                                                     & 69.72$_{4.65}$                                                     & 73.52$_{4.83}$                                                     & \textbf{-}                                                         & \textbf{-}                                                         \\ \midrule
\textbf{P2}                               & 69.04$_{12.45}$                                                     & 73.33$_{12.97}$                                                     & 71.18$_{7.16}$                                                     & 77.19$_{7.14}$                                                     & 63.91$_{4.73}$                                                     & 67.63$_{5.13}$                                                     & \textbf{80.38$_{5.81}$}                                            & \textbf{82.03$_{5.94}$}                                            \\ \midrule
\textbf{P3}                               & \textbf{77.76$_{9.52}$}                                            & \textbf{81.11$_{9.89}$}                                            & \textbf{76.52$_{7.06}$}                                            & \textbf{82.23$_{7.13}$}                                            & 63.92$_{4.88}$                                                     & 67.73$_{4.96}$                                                     & \textbf{-}                                                         & \textbf{-}                                                         \\ \midrule
\textbf{P4}                               & 67.65$_{16.54}$                                                     & 71.02$_{16.89}$                                                     & 69.23$_{8.41}$                                                     & 74.32$_{8.59}$                                                     & 68.95$_{4.91}$                                                     & 72.31$_{4.95}$                                                     & 75.84$_{4.88}$                                                     & 77.56$_{4.97}$                                                     \\ \midrule
\textbf{P5}                               & 65.18$_{17.51}$                                                     & 68.75$_{17.93}$                                                     & 69.81$_{7.86}$                                                     & 75.01$_{7.97}$                                                     & 68.44$_{4.63}$                                                     & 72.09$_{4.81}$                                                     & \textbf{-}                                                         & \textbf{-}                                                         \\ \bottomrule
\end{tabular}
}
\caption{Effect of different text prompt templates on the segmentation performance (\%, mean$_{std}$)}
\label{tab:text_ablation}
\end{table*}

\subsubsection{Experimental setup and metrics}
\noindent We performed a comprehensive comparison of segmentation quality using the initial labels derived from post-processed M2IB results, zero-shot pseudo-masks, and weakly supervised outputs on the specified testing datasets. Our zero-shot method was benchmarked against SOTA zero-shot segmentation methods, such as SaLIP \citep{aleem2024testtimeadaptationsalipcascade} and SAMAug \citep{dai2024samaugpointpromptaugmentation} and few-shot approaches, such as UniverSeg \citep{butoi2023universeguniversalmedicalimage}, ProtoSAM \citep{ayzenberg2024protosam}, and Self-Prompt-SAM \citep{wu2023self}. Additionally, we compare our weakly supervised method with nnUNet \citep{isensee2021nnu} trained on pseudo-labels without checkpoint ensembling. For weakly supervised segmentation, we trained the nnUNet \citep{isensee2021nnu} architecture for 600 epochs with 3 cycles for all datasets. The learning rate was initialized to 0.01 and we adopted a cyclical learning rate schedule as described in \citep{zhao2022efficient}, where the learning rate oscillates between a maximum and minimum value throughout each cycle. This allows the model to escape local optima and explore a wider solution space, leading to more diverse and robust predictions. We saved the last 10 checkpoints in each of the 3 cycles resulting in 30 total model checkpoints. The final segmentation result is averaged from the predictions of these 30 checkpoints and is later thresholded to create a binary mask.

As part of the ablation studies for zero-shot segmentation, we examined: \textbf{1)} the impact of fine-tuning BiomedCLIP and the choice of explainable AI (XAI) technique for saliency map generation, \textbf{2)} the influence of different text prompts on overall segmentation performance, \textbf{3)} the contribution of each model component to the final performance, and \textbf{4)} the selection of SAM pre-trained models with various visual prompting strategies. These ablation studies were conducted on the test sets of all four datasets mentioned. 

In all experiments, Dice-Sørensen Coefficient (DSC) and Normalized Surface Distance (NSD) were used as evaluation metrics. Paired-sample t-tests were also conducted to validate the observed trends, with a p-value of less than 0.05 indicating statistical significance.

\section{Results}
\subsection{Comparison with SOTA Methods}
Table \ref{tab:sota_results} shows a comparison of the proposed MedSAM-CLIPv2 with different SOTA techniques. Compared to the original MedCLIP-SAM, our approach significantly improved the average DSC from \textbf{64.54\%} to \textbf{77.61\%} and NSD from \textbf{66.10\%} to \textbf{81.56\%} in the zero-shot setting. Similarly, in the weakly supervised scenario, the average DSC increased from \textbf{70.90\%} to \textbf{82.11\%} and NSD from \textbf{73.77\%} to \textbf{87.33\%}, even surpassing weakly supervised nnUNet trained on pseudo-labels without checkpoint ensembling on average. Overall, our method significantly outperformed all zero-shot and few-shot SOTA methods across various imaging modalities/tasks ($p$ $<$ 0.05), except for Lung X-ray. However, the fully supervised methods still offer higher accuracy than those using limited resources. 

 \subsection{Effectiveness of DHN-NCE}
 The accuracy of cross-modal retrieval (text-to-image and image-to-text) for the ROCO dataset \citep{Pelka2018RadiologyOI} is shown in Table \ref{tab:cross-modal} across different losses for fine-tuning BiomedCLIP, with three pre-trained CLIP models as baselines. It can be seen that domain-specific pre-trained models performed better than CLIP, with the larger-scale pretraining offering better retrieval accuracy while the pre-trained BiomedCLIP demonstrating the highest retrieval accuracy among all pre-trained models. Fine-tuning BiomedCLIP further enhanced its performance. Specifically, BiomedCLIP fine-tuned with DHN-NCE reached \textbf{84.70\%} top-1 and \textbf{94.73\%} top-2 in image-to-text retrieval, and \textbf{85.99\%} top-1 and \textbf{95.17\%} top-2 in text-to-image retrieval, significantly outperforming other loss functions and the baseline models ($p<$ 0.01). Additionally, the benefit of fine-tuning BiomedCLIP with our DHN-NCE loss is further validated with improved segmentation quality across different tasks and image modalities in Table \ref{tab:component_ablation} and Table \ref{tab:saliency_ablation}.

\subsection{Ablation Experiments}
\subsubsection{Effect of text prompt designs}
\label{sec:text-prompt-ablation}
We conducted a series of experiments to analyze the impact of various text prompt designs on zero-shot segmentation performance. In particular, we compared six different prompt configurations: \textbf{P0} and \textbf{P1} include the class name of the object to be segmented, while \textbf{P2} and \textbf{P3} consist of longer, descriptive single prompts, and finally \textbf{P4} and \textbf{P5} are ensembles of 20 text prompts. Note that \textbf{P0}, \textbf{P2}, and \textbf{P4} are generic text prompts, while \textbf{P1}, \textbf{P3}, and \textbf{P5} are more nuanced with subtypes of the target object of interests. For example, for Breast Ultrasound, \textbf{P0} is ``\textit{breast tumor}" while \textbf{P1} can either be ``\textit{malignant breast tumor}" or ``\textit{benign breast tumor}" depending on the tumor class. For \textbf{P2}, we used one descriptive sentence, such as ``\textit{A medical breast mammogram showing a suspicious, irregularly shaped mass suggestive of a breast tumor.}" \textbf{P3}, on the other hand, includes descriptive text about a specific tumor subtype, like ``\textit{A medical breast mammogram showing an irregularly shaped, spiculated mass suggestive of a malignant breast tumor.}" \textbf{P4} and \textbf{P5} are similar to \textbf{P2} and \textbf{P3}, but they use an ensemble approach by averaging the text embeddings of 20 different prompts. Here, all descriptive clinical prompts are generated using GPT-4 \citep{achiam2023gpt}. For Lung CT, we evaluated solely on generic prompts as there is only one class available. As shown in Table \ref{tab:text_ablation}, the choice of text prompt significantly influences segmentation performance. Class-specific prompts (\textbf{P3}) generally yielded better results for smaller structures like breast and brain tumors whereas generic prompts (\textbf{P0}, \textbf{P2}) performed better for larger structures like lungs in X-ray and CT scans, where simpler, more generic descriptions allowed the model to focus on larger areas. The best prompt configuration for each task is used to generate the results presented in Table \ref{tab:sota_results}.
\begin{figure*}[t!]
    \includegraphics[scale=0.45]{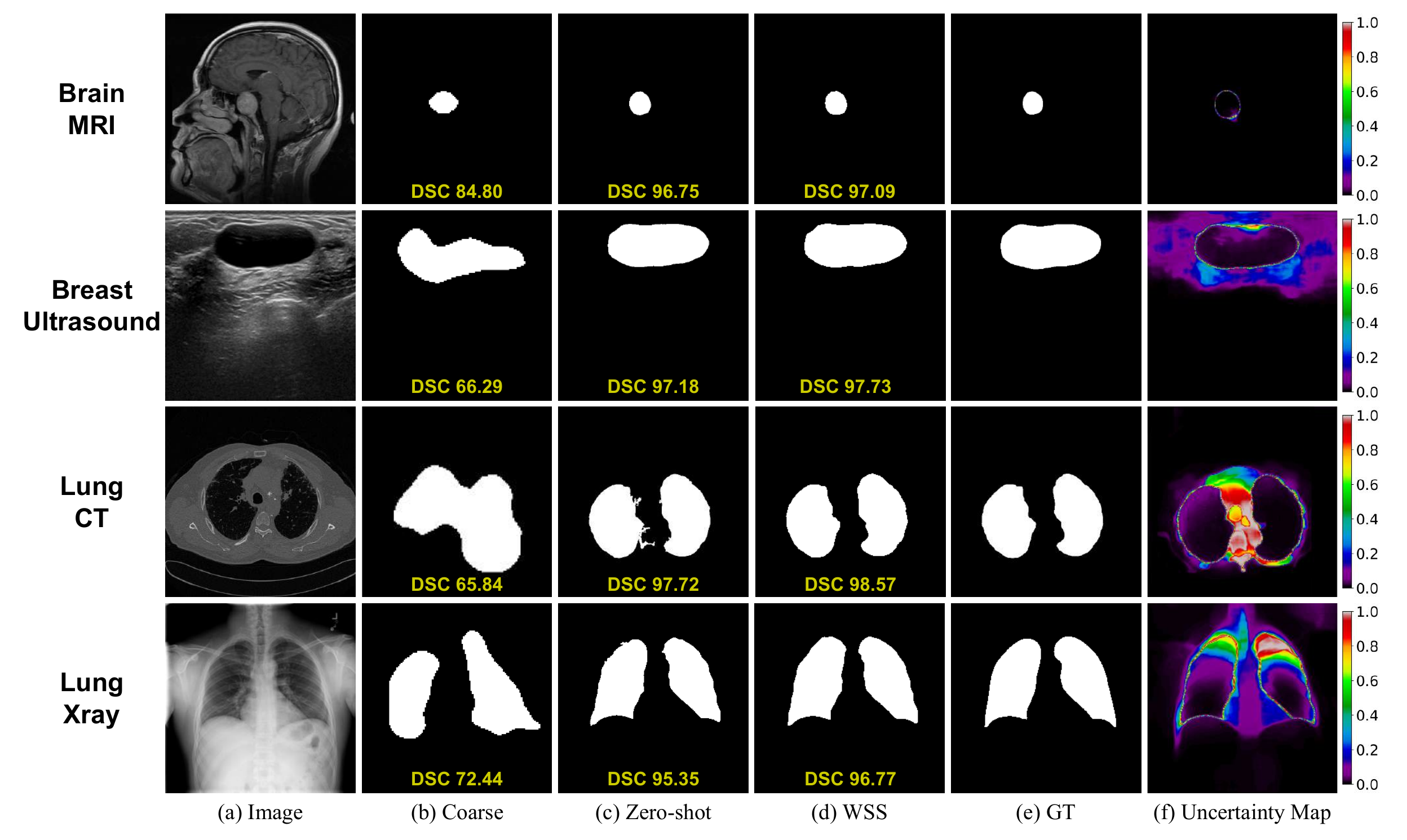}
    \caption{Qualitative comparison of segmentation results. Coarse=post-processed saliency map, WSS=Weakly Supervised Segmentation and GT=Ground Truth. The uncertainty map corresponds to the weakly supervised segmentation.}
    \label{fig:seg-illustration}
\end{figure*}
\subsubsection{Ablation Analysis of Algorithm Components}
Table \ref{tab:component_ablation} shows the contribution of each component of our framework in improving the average segmentation performance on all datasets. Starting with saliency maps generated using the M2IB, we achieved a baseline DSC of \textbf{46.23\%} and an NSD of \textbf{50.50\%}, providing an initial focus on key regions of interest. Fine-tuning BiomedCLIP with the proposed DHN-NCE loss raised the DSC to \textbf{49.10\%} and the NSD to \textbf{53.54\%}. Post-processing the saliency maps further enhanced the segmentation quality, allowing the model to better delineate foreground and background areas by refining the initial segmentation boundaries. Incorporating a connected component analysis step greatly impacted the results, increasing the DSC to \textbf{57.89\%} and the NSD to \textbf{61.54\%}, as it eliminated small, irrelevant clusters and reduced noise, improving overall precision. With the integration of SAM and the use of visual prompts, such as bounding boxes or points, our zero-shot method yielded a substantial improvement, achieving a DSC of \textbf{77.61\%} and an NSD of \textbf{81.56\%}. Finally, weakly supervised training with checkpoint ensembling further refined the segmentation quality by leveraging pseudo-labels generated from the zero-shot method. By using these pseudo-labels to fine-tune a segmentation network, we were able to reach a final DSC of \textbf{82.11\%} and an NSD of \textbf{87.33\%}.

\begin{table}[htb]
\tablestyle{-22pt}{1.1}
\addtolength{\tabcolsep}{+30pt}
\resizebox{0.5\textwidth}{!}{%
\begin{tabular}{lll}
\toprule
Method                       & \multicolumn{1}{c}{\textbf{DSC}$\uparrow$} & \multicolumn{1}{c}{\textbf{NSD}$\uparrow$} \\ \midrule
1: Saliency Maps             & 46.23$_{8.58}$                                       & 50.50$_{8.86}$                                       \\
2: + DHN-NCE Fine-tuning     & 49.10$_{8.46}$                                       & 53.54$_{8.62}$                                       \\
3: + Post-processing & 51.62$_{7.57}$                                       & 55.23$_{7.47}$                                       \\
4: + Connected Component Analysis               & 57.89$_{7.87}$                                       & 61.54$_{8.02}$                                       \\ \midrule
 \rowcolor[gray]{0.92} 5: + SAM   & 77.61$_{6.82}$                               & 81.56$_{7.00}$ \\
\rowcolor[gray]{0.83} 6: + nnUNet Ensemble & \textbf{82.11$_{8.49}$} & \textbf{87.33$_{8.46}$} \\ \bottomrule
\end{tabular}
}
\caption{Effect of different components (\%, mean$_{std}$)}
\label{tab:component_ablation}
\end{table}
\subsubsection{Impact of Saliency Maps Generation Methods}
As shown in Table \ref{tab:saliency_ablation}, M2IB achieved the highest performance across all tasks, with an average DSC of \textbf{77.61\%} and NSD of \textbf{81.56\%} when using the fine-tuned BiomedCLIP model. In both its pre-trained and fine-tuned forms, M2IB significantly outperformed gScoreCAM and GradCAM (\(p\) \(<\) 0.05). BiomedCLIP fine-tuning improved the scores across all saliency map techniques on average, with the largest gains seen in M2IB, which improved by \textbf{3.92\%} in DSC and \textbf{4.24\%} in NSD compared to its pre-trained version.
\begin{table}[htb]
\centering
\tablestyle{-22pt}{1.1}
\addtolength{\tabcolsep}{+30pt}
\resizebox{0.5\textwidth}{!}{%
\begin{tabular}{cccc}
\toprule
\multirow{2}{*}{\textbf{Model}} & \multirow{2}{*}{\textbf{Technique}} & \multicolumn{2}{c}{\textbf{All}}                                                                 \\ \cmidrule{3-4} 
                                &                                     & \textbf{DSC} $\uparrow$ & \textbf{NSD} $\uparrow$ \\ \midrule
\multirow{3}{*}{\shortstack[l]{Pre-trained \\ BiomedCLIP}}     & M2IB                                & 73.69$_{7.58}$                                                     & 77.32$_{7.43}$                                                     \\
                                & gScoreCAM                           & 58.92$_{6.67}$                                                     & 62.19$_{6.02}$                                                     \\
                                & GradCAM                             & 29.21$_{8.74}$                                                     & 31.36$_{8.44}$                                                     \\ \midrule
\multirow{3}{*}{\shortstack[l]{Fine-tuned \\ BiomedCLIP}}           & \cellcolor[gray]{0.9} M2IB                                & \cellcolor[gray]{0.9} \textbf{77.61$_{6.82}$}                                            & \cellcolor[gray]{0.9} \textbf{81.56$_{7.00}$}                                            \\
                                & gScoreCAM                           & 60.52$_{6.41}$                                                     & 63.89$_{6.39}$                                                     \\
                                & GradCAM                             & 30.11$_{8.92}$                                                     & 32.61$_{8.83}$                                                     \\ \bottomrule
\end{tabular}
}
\caption{Comparison between different Saliency Map techniques as well as the pre-trained and fine-tuned BiomedCLIP on the overall performance (\%, mean$_{std}$)}
\label{tab:saliency_ablation}
\end{table}
\begin{table*}[t!]
\centering
\tablestyle{-26pt}{1.1}
\addtolength{\tabcolsep}{+30pt}
\resizebox{\textwidth}{!}{%
\begin{tabular}{ccccccccccc}
\toprule
\multirow{2}{*}{\textbf{Model}} & \multirow{2}{*}{\textbf{Type}} & \multirow{2}{*}{\textbf{Prompts}} & \multicolumn{2}{c}{\textbf{Breast Ultrasound}}                                     & \multicolumn{2}{c}{\textbf{Brain MRI}}                                             & \multicolumn{2}{c}{\textbf{Lung X-ray}}                                            & \multicolumn{2}{c}{\textbf{Lung CT}}                                               \\ \cmidrule{4-11} 
                                &                                &                                   & \textbf{DSC} $\uparrow$ & \textbf{NSD} $\uparrow$ & \textbf{DSC} $\uparrow$ & \textbf{NSD} $\uparrow$ & \textbf{DSC} $\uparrow$ & \textbf{NSD} $\uparrow$ & \textbf{DSC} $\uparrow$ & \textbf{NSD} $\uparrow$ \\ \midrule
\multirow{3}{*}{SAM}            & \multirow{3}{*}{ViT-H}         & Points                            & 65.56$_{9.89}$                                    & 68.20$_{9.97}$                                    & 65.54$_{8.45}$                                    & 70.73$_{8.32}$                                    & \textbf{75.79$_{3.44}$}                           & \textbf{80.88$_{3.52}$}                           & 61.49$_{6.25}$                                    & 63.90$_{6.74}$                                    \\
                                &                                & BBoxes                            & \textbf{77.76$_{9.52}$}                           & \textbf{81.11$_{9.89}$}                           & \textbf{76.52$_{7.06}$}                           & \textbf{82.23$_{7.12}$}                           & 70.55$_{5.38}$                                    & 74.12$_{5.77}$                                    & \textbf{80.38$_{5.81}$}                           & \textbf{82.03$_{5.94}$}                           \\
                                &                                & Points + BBoxes                   & 74.38$_{10.57}$                                    & 79.60$_{10.62}$                                    & 75.48$_{8.66}$                                    & 80.29$_{8.64}$                                    & 73.30$_{5.94}$                                    & 79.22$_{6.12}$                                    & 62.83$_{6.72}$                                    & 64.57$_{6.99}$                                    \\ \midrule
\multirow{3}{*}{SAM-Med2D}      & \multirow{3}{*}{ViT-B}         & Points                            & 73.12$_{9.51}$                                    & 75.16$_{9.13}$                                    & 66.78$_{9.97}$                                    & 70.12$_{9.75}$                                    & 60.58$_{7.43}$                                    & 64.42$_{7.73}$                                    & 65.94$_{7.17}$                                    & 68.05$_{7.99}$                                    \\
                                &                                & BBoxes                            & 75.22$_{10.04}$                                    & 80.03$_{10.94}$                                    & 55.21$_{9.85}$                                    & 61.34$_{9.93}$                                    & 30.18$_{11.15}$                                    & 36.35$_{11.23}$                                    & 63.10$_{8.57}$                                    & 68.59$_{8.48}$                                    \\
                                &                                & Points + BBoxes                   & 74.83$_{10.78}$                                    & 79.50$_{10.12}$                                    & 67.85$_{10.96}$                                    & 72.04$_{10.45}$                                    & 37.23$_{8.69}$                                    & 44.90$_{9.37}$                                    & 71.22$_{8.09}$                                    & 78.05$_{8.11}$
                                    \\ \midrule
MedSAM                          & ViT-B                          & BBoxes                            & 63.50$_{11.42}$                                    & 68.11$_{11.25}$                                    & 67.68$_{12.75}$                                    & 73.89$_{12.67}$                                    & 73.03$_{6.03}$                                    & 76.23$_{6.02}$                                    & 62.14$_{7.80}$                                    & 65.00$_{7.11}$  
                                  \\ \bottomrule
\end{tabular}
}
\caption{Comparison between different SAM pre-trained models and visual prompting techniques (\%, mean$_{std}$)}
\label{tab:prompt_sam_ablation}
\end{table*}

\subsubsection{Comparison of Visual Prompts for SAM}
Table \ref{tab:prompt_sam_ablation} compares different SAM models and visual prompting techniques. We see that bounding boxes generally provided the best segmentation performance, except in Lung X-rays, where adding point prompts enhanced results. On the other hand, point prompts alone performed worse except in certain tasks, such as Lung X-ray (\textbf{75.79\%} DSC, \textbf{80.88\%} NSD). In addition, the comparison of SAM, MedSAM, and SAM-Med2D demonstrates that SAM, despite not being pre-trained on medical data, performs well with bounding box prompts, achieving high scores in most modalities/tasks, including Lung CT. SAM-Med2D excels in fine-scaled segmentation, but struggles with larger structures, like lung lobes, where MedSAM performs better. The superior performance of SAM may be attributed to its use of a larger model architecture (ViT-H) compared to MedSAM and SAM-Med2D, which only offer ViT-B configurations.

\subsection{Qualitative Segmentation Results}
Lastly, we present qualitative segmentation results across the four imaging modalities evaluated for our proposed method in Fig. \ref{fig:seg-illustration}. Our proposed MedCLIP-SAMv2 consistently produced high-quality segmentation masks in weakly supervised settings. For all datasets except Brain MRI, the initial coarse segmentation was suboptimal. However, it provided a sufficient starting point for the zero-shot approach to refine coarse activation maps. For breast and brain tumors, the zero-shot results were notably better than those for Lung CT and Lung X-ray. In Lung CT, the primary challenge for the algorithm was distinguishing between the two lobes. The post-processed results showed one large, connected contour in the center. The zero-shot refinement slightly separated these two regions, though some artifacts persisted. However, the weakly supervised training effectively corrected these false activations, producing a high-quality segmentation map. For Lung X-ray, while the weakly supervised training improved upon the less precise zero-shot masks, the improvement was not as substantial as with Lung CT. Furthermore, we also included uncertainty maps for all predictions. For Brain MRI, high uncertainty was observed only at the edges of the segmentation, which is typical. For Breast Ultrasound, high uncertainty was observed at the borders of the segmentation, while the surrounding area outside the borders showed low uncertainty. In contrast, for Lung X-rays, slight uncertainty appeared in the center of the mask, increasing towards the edges. In the case of Lung CT, high uncertainty was observed both at the edges and in the center of the lung lobes. This was largely due to the artifacts present in the zero-shot pseudo-labels.

\section{Discussion}
The proposed MedCLIP-SAMv2 framework demonstrates superior performance in zero-shot and weakly supervised medical image segmentation tasks than SOTA methods and the original MedCLIP-SAM method \citep{koleilat2024medclipsambridgingtextimage} across four critical medical imaging modalities (CT, MRI, Ultrasound, and X-ray). By leveraging BiomedCLIP and SAM with text and visual prompts, our method exhibits robust domain and task generalization, excelling in complex tasks, such as brain and breast tumor segmentation, where smaller and intricate anatomical details pose challenges in typical segmentation tasks. Our approach notably surpasses other SOTA zero-shot and few-shot methods, especially in difficult segmentation scenarios (see Table \ref{tab:sota_results}). Recent methods like \citep{ding2022decouplingzeroshotsemanticsegmentation} have demonstrated the potential of CLIP for zero-shot segmentation by decoupling the pixel-level and image-level classification tasks in natural vision applications. However, such methods require fully supervised segmentation ground truths, limiting their application in settings where labels are scarce or noisy, like medical image segmentation. In contrast, MedCLIP-SAMv2 bypasses this requirement and operates without relying on segmentation labels during training, offering a more scalable approach for medical imaging, particularly in weakly supervised settings.

\begin{figure}
    \centering
    \includegraphics[width=0.5\textwidth]{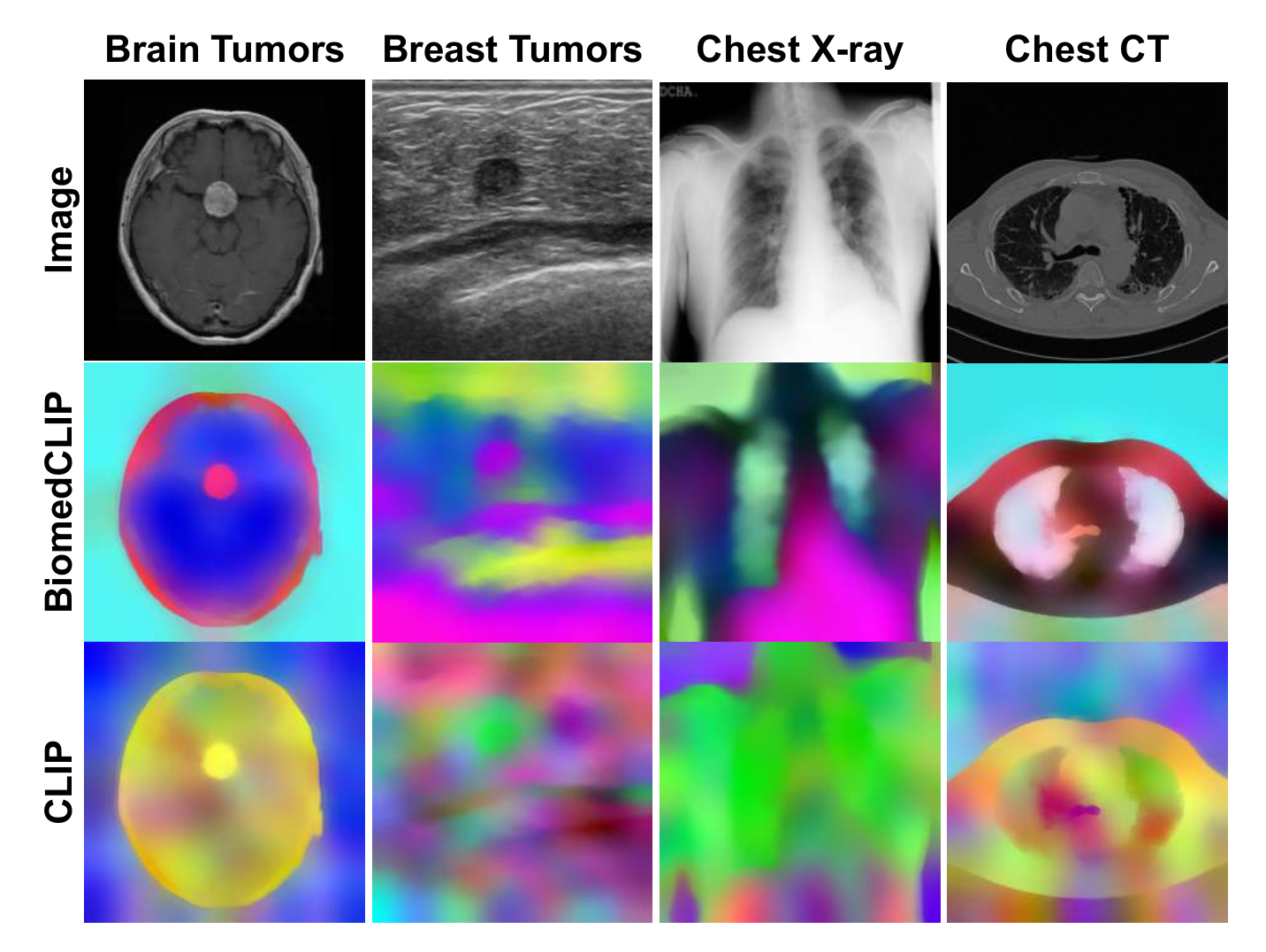}
    \caption{Diagram showing upsampled feature representations from the last transformer layer of CLIP and BiomedCLIP. Feature Maps were upsampled using FeatUp \citep{fu2024featup} for visualization purposes.}
    \label{fig:CLIP_Features}
\end{figure}

Compared with the original MedSAM-CLIP, the component updates in MedSAM-CLIPv2 have greatly contributed to the performance improvement. One of the key strengths of our framework lies in the integration of M2IB for radiological tasks, which effectively extracts meaningful information from medical images and texts, enhancing segmentation performance. The introduction of the DHN-NCE loss played a crucial role in fine-tuning BiomedCLIP, enabling the model to focus on challenging details while maintaining high performance across all tasks and modalities. Importantly, the combination of M2IB and DHN-NCE allowed the model to generate coarse segmentation masks that are later refined via SAM in a zero-shot setting (see Table \ref{tab:saliency_ablation}), proving the versatility of the method without the need for ground truth annotations. Finally, the effectiveness of prompt design was another critical insight. Contextually rich, descriptive prompts yielded better results in complex tasks like tumor segmentation, where finer anatomical understanding is required. Conversely, more generic prompts sufficed for simpler tasks like lung segmentation, where larger, distinct structures allowed the model to achieve strong performance with less specific guidance. This insight suggests the importance of tailoring the text prompts in visional language models for specific radiological tasks. This contrasts with findings from other studies that used the frozen BiomedCLIP encoder with an added decoder head for segmentation transfer learning, where text prompts had little impact on segmentation quality \citep{poudel2023exploring}. The choice of BiomedCLIP over CLIP also facilitates the success of our method. Figure \ref{fig:CLIP_Features} shows the latent representations produced by CLIP and BiomedCLIP (both utilizing the same architecture i.e. ViT-B/16) of sample medical images. The latter shows that the BiomedCLIP model learns to encode meaningful latent representations of salient regions within medical scans from only natural language supervision, facilitating its ability to highlight disease-relevant regions across various modalities compared to CLIP where the subtle visual cues found in medical images are not sufficiently captured or distinguished. 

Our framework's ability to operate in a weakly supervised paradigm further strengthens its potential clinical applicability. By using pseudo-labels from zero-shot segmentation to fine-tune the model, we observed notable improvements, particularly in lung CT segmentation, where the combination of zero-shot labels and weak supervision generated significant accuracy gains. To the best of our knowledge, we are the first to integrate uncertainty estimation through nnUNet with checkpoint ensembling by training on pseudo-segmentation data, providing a robust method for enhancing segmentation quality while offering insights into prediction confidence for potential end users. Uncertainty measures are essential in clinical adoption, as they help identify regions, where the model's predictions are less certain, enabling clinicians to focus on areas that may require further examination or validation.  

Despite the original SAM model not being pre-trained on medical images, it showed strong performance in zero-shot settings, outperforming MedSAM and SAM-Med2D when provided with imperfect visual prompts like points and/or bounding boxes. This underscores the robustness of SAM to suboptimal input conditions as highlighted by \citep{HUANG2024103061}. Specifically, this can be seen in Fig. \ref{fig:seg-illustration}, where even coarse segmentations can be refined using both zero-shot and weakly supervised methods. Looking ahead, our future work will focus on extending our framework to handle 3D medical data, a crucial step in advancing the segmentation of volumetric imaging modalities like MRI and CT. Incorporating 3D models will enable our framework to better capture complex anatomical structures, further enhancing its clinical utility. Overall, our findings show that MedCLIP-SAMv2, with its integrated components, marks a significant step forward in the development of universal, interactive medical image segmentation. The framework's adaptability across different tasks and its ability to operate with minimal labeled data emphasize its potential for clinical adoption, particularly in resource-constrained settings. For our exploration, we focused on radiological tasks, with image modalities having more distinct characteristics than natural images. In the future, we will further incorporate and assess the performance of photograph-based biomedical images, such as histopathological images and surgical video with our proposed framework.

\section{Conclusion}
We presented MedCLIP-SAMv2, an upgraded version of the original MedCLIP-SAM framework, significantly improving segmentation performance with minimal supervision across CT, X-ray, Ultrasound, and MRI. By introducing the novel DHN-NCE loss for fine-tuning BiomedCLIP and leveraging SAM, our model achieved enhanced accuracy, particularly in complex tasks. MedCLIP-SAMv2 outperforms its predecessor through superior generalization and refined segmentation, demonstrating strong potential for clinical use in data-limited environments.
\section*{Acknowledgment}
We acknowledge the support of the Natural Sciences and Engineering Research Council of Canada (NSERC).
\bibliographystyle{elsarticle-harv}
\bibliography{references.bib}

\end{document}